\documentclass[10pt,twocolumn,letterpaper]{article}

\usepackage[pagenumbers]{cvpr} 

\usepackage{amsmath,amsfonts}
\usepackage{algorithmic}
\usepackage{algorithm}
\usepackage{array}
\usepackage{textcomp}
\usepackage{stfloats}
\usepackage{url}
\usepackage{verbatim}
\usepackage{graphicx}
\usepackage{cite}
\usepackage{booktabs}
\usepackage{color}
\usepackage{indentfirst}
\usepackage{multirow}
\usepackage{threeparttable}
\usepackage{subcaption}
\usepackage{listings}

\usepackage[accsupp]{axessibility}

\usepackage[dvipsnames]{xcolor}

\definecolor{cvprblue}{rgb}{0.21,0.49,0.74}
\usepackage[pagebackref,breaklinks,colorlinks,citecolor=cvprblue]{hyperref}

\def\paperID{10584} 
\def\confName{CVPR}
\def\confYear{2024}

\title{X-3D: Explicit 3D Structure Modeling for Point Cloud Recognition }
\author{Shuofeng Sun\textsuperscript{1}, Yongming Rao\textsuperscript{2}, Jiwen Lu\textsuperscript{3}, Haibin Yan\textsuperscript{1}\thanks{Corresponding author}\\
Beijing University of Posts and Telecommunications\textsuperscript{1}, Tencent\textsuperscript{2}, Tsinghua University\textsuperscript{3} \\
}

\begin{document}
\maketitle
\begin{abstract}
Numerous prior studies predominantly emphasize constructing relation vectors for individual neighborhood points and generating dynamic kernels for each vector and embedding these into high-dimensional spaces to capture implicit local structures.  However, we contend that such implicit high-dimensional structure modeling approch inadequately represents the local geometric structure of point clouds due to the absence of explicit structural information.  Hence, we introduce X-3D, an explicit 3D structure modeling approach.  X-3D functions by capturing the explicit local structural information within the input 3D space and employing it to produce dynamic kernels with shared weights for all neighborhood points within the current local region.  This modeling approach introduces effective geometric prior and significantly diminishes the disparity between the local structure of the embedding space and the original input point cloud, thereby improving the extraction of local features. Experiments show that our method can be used on a variety of methods and achieves state-of-the-art performance on segmentation, classification, detection tasks with lower extra computational cost, such as \textbf{90.7\%} on ScanObjectNN for classification, \textbf{79.2\%} on S3DIS 6 fold and \textbf{74.3\%} on S3DIS Area 5 for segmentation, \textbf{76.3\%} on ScanNetV2 for segmentation and \textbf{64.5\%} mAP$_{25}$, \textbf{46.9\%} mAP$_{50}$ on SUN RGB-D and \textbf{69.0\%} mAP$_{25}$, \textbf{51.1\%} mAP$_{50}$ on ScanNetV2. Our code is available at 
\href{https://github.com/sunshuofeng/X-3D}{https://github.com/sunshuofeng/X-3D}.
\end{abstract}    
\section{Introduction}
\label{sec:intro}

Using deep learning for point cloud analysis has garnered significant attention in research.  However, the irregularity of point cloud data poses challenges for directly applying conventional convolutional methods.

\begin{figure}[!t]
 \begin{subfigure}[t]{0.5\textwidth}
           \centering
           \includegraphics[width=\textwidth]{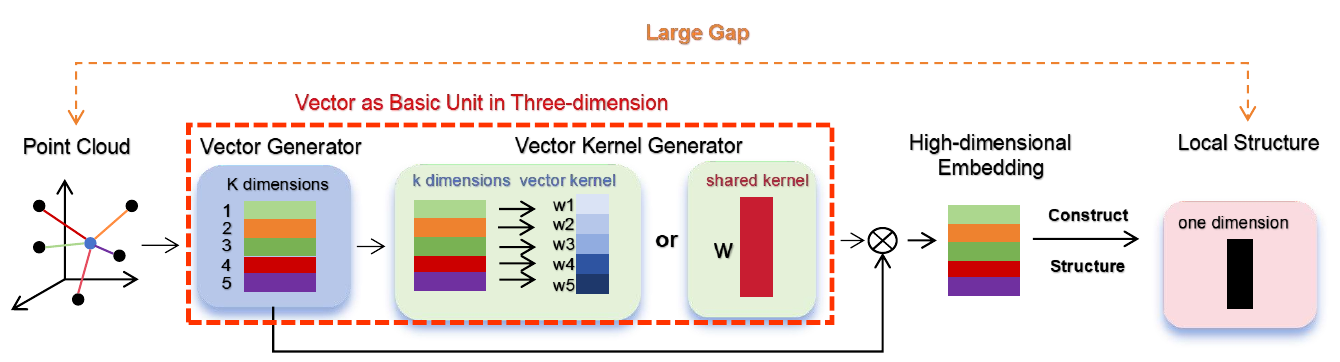}
            \caption{Implict High-dimensional Structure Modeling}
            \label{fig:diagm_ihsm}
\end{subfigure}
\begin{subfigure}[t]{0.5\textwidth}
        \centering
        \includegraphics[width=\textwidth]{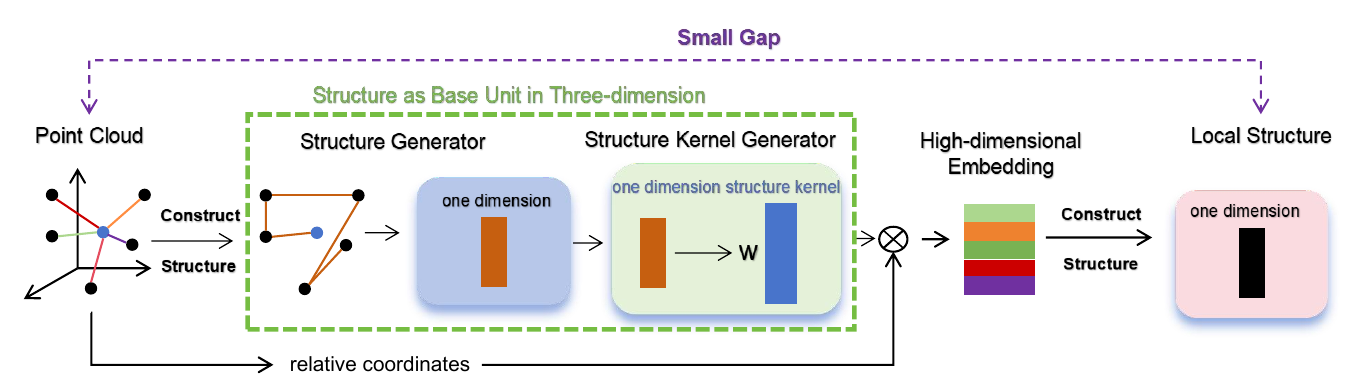}

        \caption{Explicit Three-dimensional Structure Modeling}
        \label{fig:diagm_x3d}
\end{subfigure}
\caption{Illustration of the different design paradigms. \textbf{Implict High-dimensional Structure Modeling (IHSM)}. Most of  existing work can be classified into this design paradigm, and the focus of modeling usually lies in how to construct relation vectors for each neighborhood point or how to generate a dynamic kernel for each vector and then embed the relation vectors to the high-dimension space to capture the implicit local structure.   \textbf{Explicit 3D Structure Modeling}. The difference is that the basic unit of our modeling is the structure, and we directly build the explicit geometric structure for the local neighborhood in the input space, and generate the dynamic kernel which shares weights for all neighborhood points within the current local region through the geometric structure. By this explicit introduction of structural information into the embedding space, we greatly reduce the gap between the local structure captured by the embedding space and the original input point cloud.
}\label{fig:overview}
\vspace{-1.1em}
\end{figure}

In order to solve this problem, the existing methods~\cite{pointnet++,pointnetxt,relation-shape,hu2023gam,lin2023meta,kpconv,zhao2021point} first construct the local neighborhood by kNN or ball query, and then design the relationship vector to represent the geometric relationship between each neighborhood point and the center point. For example, PointNet++~\cite{pointnet++}, PointNeXt~\cite{pointnetxt}, and other methods directly use the relative coordinate difference as the relationship vector. Methods such as RandLA~\cite{hu2020randla} and RSConv~\cite{relation-shape} supplement the relation vectors with additional Euclidean distances and absolute coordinates. Then most methods use shared-weight MLP to embed the relation vector into the high-dimensional feature space. However, in order to flexibly handle different relation vectors, some methods~\cite{kpconv,relation-shape} dynamically generate the corresponding vector kernel for each relation vector. 
Finally, the implicit local structure is obtained by applying a symmetric aggregation function like attention mechanism~\cite{zhao2021point,lai2022stratified} or max-pooling to all embedding vectors.
We refer to the above modeling process as implicit high-dimensional structure modeling, where the basic units are vectors, and then the local structure is only implicitly captured in high-dimensional space.

\begin{figure}[!t]
    \centering
    \begin{subfigure}[t]{0.5\textwidth}
           \centering
           \includegraphics[width=\textwidth]{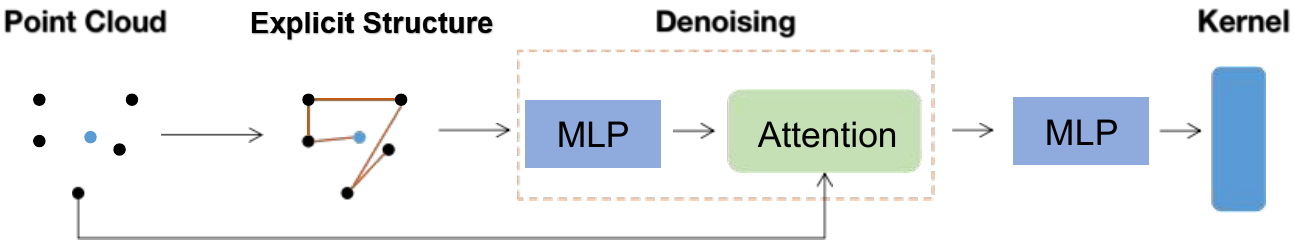}
            \caption{Structure Kernel Generator}
            \label{fig:network1}
    \end{subfigure}
    \begin{subfigure}[t]{0.5\textwidth}
            \centering
            \includegraphics[width=\textwidth]{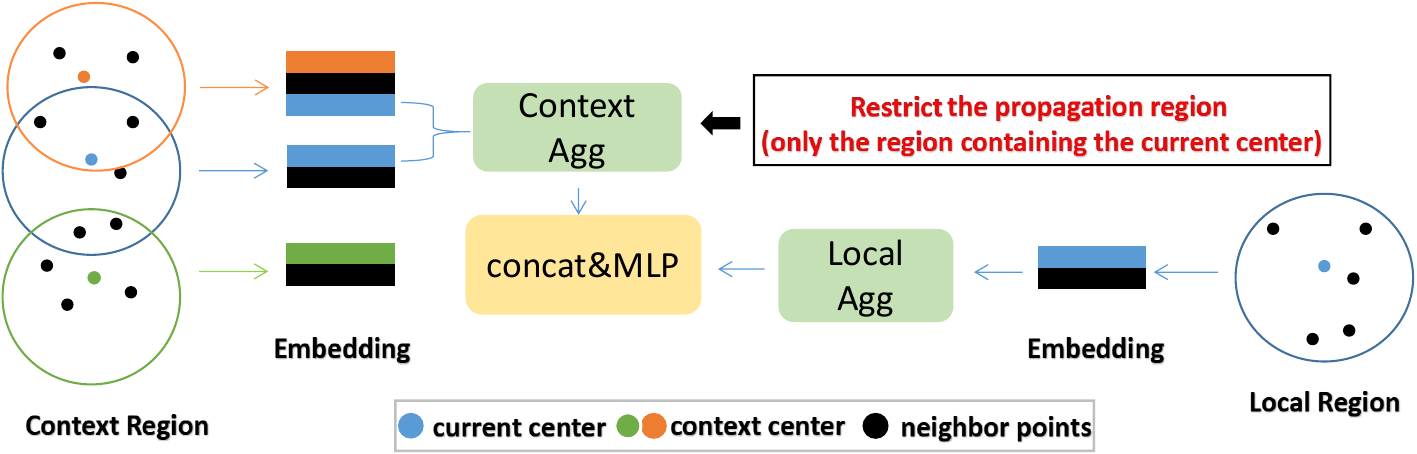}

            \caption{Neighborhood Context Propagation}
            \label{fig:network2}
    \end{subfigure}
    
    \caption{Illustration of X-3D. \textbf{(a)} first constructs the explicit local structure from the original input space, then reduces the influence of noise points on the local structure by cross attention, and finally generates the structure kernel by MLP. \textbf{(b)} avoids the influence caused by random neighborhood selection by propagating the neighborhood context,. Furthermore, by limiting the scope of dynamic context propagation, it ensures that the explicit local structure does not conflict}
    \label{fig:network}
    \vspace{-1.1em}
\end{figure}

However, there may be some problems with the above paradigm. First, local structure is implicitly captured in high dimensions which may lead to the loss of some detailed information and a large gap with the local structure of the original data. Further, point clouds are typical manifold data in the non-Euclidean space  and thus relations vectors in the Euclidean space
may provide inaccurate geometric information (e.g., Euclidean distances are very close, and geodesic distances are very far). 
To address this issue, we propose X-3D, an explicit 3D structure modeling paradigm, which is shown in Figure~\ref{fig:overview}.  X-3D directly constructs and represents the local structure in the original 3D input space, generating structure kernels dynamically which is shown in Figure~\ref{fig:network}. Each relation vector is embedded using a dynamically generated kernel specific to its local region. As a result, the output embedding vector is highly correlated with the 3D local structure it represents. This significantly reduces the gap between the embedding space and the original input space's local structure, enabling more effective extraction of local features

Furthermore, to improve X-3D's performance, we focus on improving the neighborhood quality through denoising and neighborhood context propagation. Initially, denoising involves removing outliers from the explicit structure by employing a cross-attention mechanism between neighborhood points and the explicit local structure. Subsequently, to refine the neighborhood representation, we dynamically propagate neighborhood context information. However, to avoid conflating distant geometric structures and conflicting with local information, the scope of neighborhood context propagation is restricted.

Our contributions can be summarized as follows:
\begin{itemize}[leftmargin=2.1em]

 \item  We propose an explicit 3D structure modeling approach.
 \item  We propose denoising and neighborhood context propagation to further improve the performance.
 \item  We conduct an in-depth analysis from the viewpoint of manifold learning.
 \item  Our method can be embedded into other models and reach state-of-the-art performance.
\end{itemize}


\section{Related Work}
\noindent\textbf{Voxel-based Methods:} Due to the irregularity of the point cloud, many works~\cite{zhou2018voxelnet,deng2021voxel,wu20153d,mink,yan2018second} voxelize the input point cloud, transform it into a regular grid, and use 3D convolution for processing. However it is easy to miss the detailed information of the original point cloud.

\noindent\textbf{Point-based Methods:} Some researchers perform feature extraction directly on the raw point cloud. For example, PointNet++~\cite{pointnet++} constructed a neighborhood, then extracts each neighborhood point feature, and finally captures the implicit local structure of the neighborhood through a symmetric aggregation function. The subsequent work~\cite{repsurf,relation-shape,hu2023gam} designed a rich relationship vector to describe the neighborhood point features based on the PointNet++ design paradigm, and graph-based work~\cite{dgcnn,deepgcns,ecc} designed different edge descriptors to extract the neighborhood features, and dynamic weight (kernel)-based ~\cite{hu2020randla,kpconv,relation-shape} dynamically generated the corresponding kernel or weight for each neighborhood point.

\noindent\textbf{Manifold Learning:} Manifold data lies in the non-Euclidean space and has complex local structure. Simple Euclidean metric is difficult to measure the geometric properties of manifold data and easily leads to capture the wrong geometric structure. ISOMAP captures the manifold properties by computing geodesic distances instead of Euclidean metrics and minimizing point distances across all input and embedding spaces, while methods such as LLE~\cite{roweis2000nonlinear}, LTSA~\cite{zhang2004principal} captures the correct manifold surface by minimizing the local structure of the original input and embedding spaces. Point clouds are typical manifold data, and simply using relative coordinates to represent the geometry is not enough. 
Therefore, we propose to explicitly exploit the local structure of the input space.

 

\section{Background}
\label{subsec:3.1}
In this part, we introduce implict high-dimensional structure modeling (IHSM) in detail.
Most existing methods~\cite{pointnet++,lin2023meta,hu2020randla,kpconv,hu2023gam,repsurf,relation-shape} can be considered as this paradigm, where there are two main key designs. The first key design is how to design the relation vectors in the original input space, and the second key design is how to design the corresponding dynamic kernel for each relation vector and embed each vector to the high-dimension space. It should be noted that the shared-weight MLP can be considered as a special case of the second one.

A point cloud can be represented as a set of $N$ points, we use $\mathbf{p}\in R^{N\times 3}$ to represent the coordinates of the points and $\mathbf{f}\in R^{N\times C}$ represents the features of the points, where $C$ is the feature dimension of the points. For the $i$-th center point $\mathbf{p}_i$, we define its neighborhood points as $\mathbf{p}_{i,j}$, where $j=1,2...k$, $k$ is the neighborhood number, and then the relation vector between the neighborhood points and the center point can be written as $V_{i,j}$. 

Many works focus on how to extract local structures by constructing appropriate relation vectors, for example, PointNet++~\cite{pointnet++}, PointMetaBase~\cite{lin2023meta} and others simply treat relative coordinate differences as relation vectors, while RandLA~\cite{hu2020randla} adds absolute coordinates and Euclidean distances. GAM~\cite{hu2023gam} and RepSurf~\cite{repsurf} consider depth gradient information and curvature, respectively. In particular, RepSurf explicitly incorporates local structure during curvature construction, primarily performed during preprocessing. However, as the down-sampling rate and local neighborhood half-size increase, RepSurf struggles to accurately portray the region's local structure. Consequently, it tends to describe solely point and vector information, leading to its attribution to IHSM.

Then for each vector, some works generate its corresponding dynamic vector kernel,  which can be written as:
\begin{equation}
   \mathbf{W}_{i,j}=\theta(V_{i,j})
\end{equation}
For example, RSConv~\cite{relation-shape} simply uses MLP to generate dynamic kernels based on relation vectors and KPConv~\cite{kpconv} dynamically generates vector kernels by calculating the geometric relationship between each neighborhood point and the predefined kernel points.~In particular, for shared-weight MLP, it can be rewritten as:
\begin{equation}
   \mathbf{W}_{i,j}=\theta
\end{equation}
Then, the process the vector modeling can be written as
\begin{equation}
    \hat{\mathbf{f}}_{i,j}=\mathcal{M}(\mathbf{W}_{i,j},\mathbf{f}_{i,j},V_{i,j})
\end{equation}
where $\mathcal{M}$ represents how to update neighborhood features according to dynamic vector kernel and relation vector.

Finally, the implicit local structure is obtained by aggregating the modeled relation vectors in the neighborhood, which can be written as:
\begin{equation}
\label{eq:ls_vector}
    LS_i=\mathop{Agg}\limits _{j=1..k}(\hat{\mathbf{f}}_{i,j})=\mathop{Agg}\limits _{j=1..k}(\mathcal{M}(\theta(V_{i,j}),\mathbf{f}_{i,j},V_{i,j}))
\end{equation}

\begin{figure}[!t]
    \centering
    \begin{subfigure}[t]{0.15\textwidth}
           \centering
           \includegraphics[width=\textwidth]{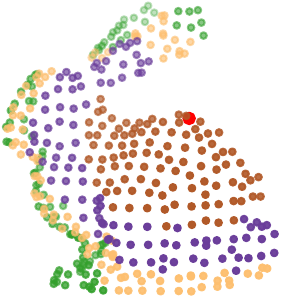}
            \caption{Geodesic Distance}
            \label{fig:origin_pc}
    \end{subfigure}
    \begin{subfigure}[t]{0.15\textwidth}
            \centering
            \includegraphics[width=\textwidth]{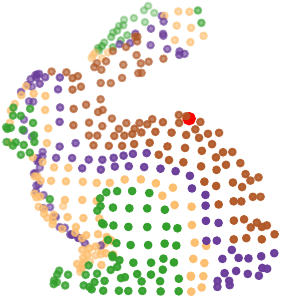}

            \caption{IHSM}

            \label{fig:geo_pc}
    \end{subfigure}
    \begin{subfigure}[t]{0.15\textwidth}
            \centering
            \includegraphics[width=\textwidth]{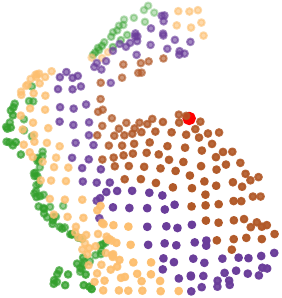}
            \caption{Ours}
            \label{fig:baseline_pc}
    \end{subfigure}
    \caption{Examples of local structures captured by different methods. \textbf{(a)} captures local structure by computing geodesic distances on the original input point cloud. \textbf{(b)}  captures the implicit local structure through implicit high-dimensional structure modeling, and it can be seen that there is still a certain difference from the local structure of the original space represented by the geodesic distance. \textbf{(c)} captures the local structure through X-3D, and it can be seen that the local structure difference from the original space represented by the geodesic distance is small, and the representation in the edge part is more reasonable. }
    \label{fig:example_fc}
    \vspace{-1.1em}
\end{figure}

We can see that the basic unit that captures implicit local structure is the vector. Specifically, both the input and the corresponding kernel only consider the relation vector and ignore the overall structure. Therefore, many details are missed in the transformation from vector to structure in high-dimensional space, which may lead to a large gap with the local structure in the input point cloud data and increase the difficulty of model learning (An example in Figure~\ref{fig:example_fc}).

\section{Proposed Method}
To solve the problem of IHSM, we change the focus of design from vectors to structures and propose explicit 3D structure modeling. In simple terms, we capture explicit local structure directly in the raw input point cloud and dynamically generate structure kernels based on it, which can be written as:
\begin{equation}
   \hat{\mathbf{W}_{i}}=\theta(ES_i)
\end{equation}
where $ES_i$ represents the explicit structure of the neighborhood to the $i$-th center point. Then the neighborhood point features are updated by the following formula:
\begin{equation}
\label{eq:agg}
    \hat{\mathbf{f}}_{i,j}=\mathcal{M}(\hat{\mathbf{W}_{i}},\mathbf{f}_{i,j},\mathbf{p}_i-\mathbf{p}_{i,j})
\end{equation}

Finally, the capture of local structure in the embedding space can be written as:
\begin{equation}
\label{eq:ls_structure}
    LS_i=\mathop{Agg}\limits _{j=1..k}(\hat{\mathbf{f}}_{i,j})=\mathop{Agg}\limits _{j=1..k}(\mathcal{M}(\theta(ES_i),\mathbf{f}_{i,j},\mathbf{p}_i-\mathbf{p}_{i,j}))
\end{equation}
We can see that the similarity of (\ref{eq:ls_vector}) and (\ref{eq:ls_structure}) effectively indicates the applicability of X-3D, so the proposed approach can be easily embedded into most of the existing models and effectively improve their performance.

In the following part, we will go through the various parts of X-3D in detail.

\subsection{Explict Structure}
 For the explicit structure $ES_i$, explicitly capturing local structure has been deeply studied in previous machine learning methods, such as LLE~\cite{roweis2000nonlinear}, which takes the linear reconstruction relationship \textbf{(LR)} between the center point and the neighborhood points as the local structure, PointHop~\cite{zhang2020pointhop}, which divides the point cloud neighborhood into fixed octuples and extracts the centroid coordinates of each partition \textbf{(PH)},  and some methods~\cite{spg,pca} utilize neighborhood \textbf{PCA} to combine eigenvector and eigenvalue to obtain robust structural descriptors. 

We conducted a large number of experiments to explore which $ES_i$ has better performance. Through extensive experiments, we found that effective explicit local structures $ES_i$ should have the following properties:
\begin{itemize}[leftmargin=0.63cm]
\item  \textbf{Symmetry:} For a neighborhood of $k$ points, the 1D $ES_i$ should be symmetric, and the order of the points does not affect the capture of the explicit structure.
\item  \textbf{Global Shape:} $ES_i$ should be able to describe the global shape of the local region, which provides a good geometric prior for tasks such as segmentation.
\item  \textbf{Local Details:} $ES_i$ should simply describe the local details of the  current local region, which contributes to the correct geometric correlation of the points in the current neighborhood.
\item \textbf{Explicit information:} $ES_i$ should be explicitly derived from the data rather than  implicitly learned by the model
\end{itemize}

\hspace*{\fill}

In Table~\ref{tab:structure}, we evaluated different explicit local structures to underscore the significance of these properties. Results indicate that \textbf{IS} exhibits lower performance on GAP and OA due to a lack of explicit structural information. Similarly, \textbf{LR} lacks symmetry, affecting its generalization performance, particularly in the presence of disorder within the point cloud, resulting in lower overall accuracy (OA). Considering the higher performance and comprehensive properties of PointHop, we default to using it as the explicit local structure.
The process of PointHop can be written as:
 \begin{equation}
 \label{eq:pointhop}
    \overline{\mathbf{p}}_{i,c}=\frac{1}{K_{i,c}}\sum_{j=1}^kt_{i,j,c}\mathbf{p}_{i,j}, c=1,...8
\end{equation}
\begin{equation}
    {ES}_{i}=Concat([\overline{\mathbf{p}}_{i,1},...\overline{\mathbf{p}}_{i,8}])
\end{equation}
where $\overline{\mathbf{p}}_{i,c}$ represents the centroid coordinates of the $c$-th partition of the $i$-th point, and $K_{i,c}$ represents the number of neighborhood points of the partition. $t_{i,j,c}$ represents whether the $j$-th neighborhood point of the $i$-th center point belongs to the $c$-th partition,which can be written as:
\begin{equation}
t_{i,j,c}=\left\{
\begin{aligned}
    1&,&\mathbf{p}_{i,j} \in \xi_c\\
    0&,&\mathbf{p}_{i,j} \notin \xi_c\\
\end{aligned}
\right.
\end{equation}
where $\xi_c$ represents the $c$-th partition.

\subsection{Structure Kernel Generator.} 
Having obtained the explicit local structure $ES_i$, we use it to dynamically generate structure kernels.
First, we further extract local structural information by:
\begin{equation}
    {F}_{i}=MLP({ES}_{i})
\end{equation}

Then  we simply define noisy points as those that ``differ significantly from the local structure", therefore we  calculate the correlation between each point and local structure information, and perform adaptive fusion to eliminate the influence of noise points:
\begin{equation}
    s_{i,j}=Softmax(MLP(\mathbf{p}_{i,j})^T{F}_{i})
\end{equation}
\begin{equation}
    {F}_{i}={F}_{i}+\sum_{j=1}^Ks_{i,j}MLP(\mathbf{p}_{i,j})
\end{equation}

Then we expect the model to learn the intrinsic geometric properties of the point cloud surface through the local geometric structure, capturing the correct geometric correlations. To do this, we use the geometric structure descriptor to dynamically generate kernels.
\begin{equation}
    \hat{\mathbf{W}_{i}}=MLP(F_{i})
\end{equation}

Then, in the process of updating the neighborhood point features,  based on the practice of PointMetaBase, we explicitly add the dynamic position embedding to the neighborhood point features to provide the correct geometric correlation.
\begin{equation}
\label{eq:update}
    \mathcal{M}(\hat{\mathbf{W}_{i}},\mathbf{f}_{i,j},\mathbf{p}_i-\mathbf{p}_{i,j})=\hat{\mathbf{W}_{i}}(\mathbf{p}_i-\mathbf{p}_{i,j})+MLP(\mathbf{f}_{i,j})
\end{equation}

\begin{table}[!t]
\centering
\caption{We test the effect of different local structures on ScanObjectNN and report \textbf{OA} (overall accuracy). We additionally report the Euclidean distance (\textbf{GAP}) between the local structure of the embedding space and the local structure of the input space. To test the need for explicit structure, we add ablation experiments with implicit structure (\textbf{IS}) learned through the network. }
\label{tab:structure}

\resizebox{\linewidth}{!}{
 \setlength{\tabcolsep}{2.5pt}
\begin{tabular}{c|cccc|cc}
\toprule
&Symmetry&Glocal Shape&explicit&Local Details&GAP&OA  \\ %
\midrule
LR&&&\checkmark&\checkmark&1.48&87.2\\
IS&\checkmark&\checkmark&&\checkmark&5.37&87.6\\
PH&\checkmark&\checkmark&\checkmark&\checkmark&\textbf{1.36}&\textbf{88.8}\\
PCA&\checkmark&\checkmark&\checkmark&&1.53&88.6\\
\bottomrule
\end{tabular}
}
\end{table}

\subsection{Neighborhood Context Propagation.} 
In order to reduce the influence caused by random neighborhood point selection, we dynamically propagate the neighborhood context to supplement the complete local region information. Some previous works~\cite{hu2020randla,huang2023lcpformer,lai2022stratified} such as RandLA~\cite{hu2020randla} by multiple local feature aggregation, LCPFormer~\cite{huang2023lcpformer} by collecting features of neighborhood points in different neighborhoods to propagate local context.  These practices often expand the receptive field and fuse local structures that are far away. 

Our method generates the dynamic kernel from the explicit geometric structure of the current neighborhood. However, distant local structures might reside on different underlying manifolds, causing conflicts in geometric information. To address this, we limit context propagation to consider solely the local neighborhoods close in geometric properties to the center point, minimizing conflicts.

\begin{table*}[tb]
\center
\caption{Semantic segmentation results on the S3DIS~\cite{s3dis} 6-Fold and Area-5 dataset and the ScanNetV2~\cite{scannet} validation set. Our X-3D can effectively improve its performance on the state-of-the-art large models as well as the small-scale models, and reach the new state-of-the-art (\textbf{rank1 without extra training data}) with only a small computational overhead.
}
\label{tab:segmentation}
\resizebox{\linewidth}{!}{
\begin{tabular}{c|ccc|ccc|c|cc}
\toprule
\multirow{2}{*}{Method}&\multicolumn{3}{c|}{\textbf{S3DIS 6-fold}}&\multicolumn{3}{c|}{\textbf{S3DIS Area-5}}&\textbf{ScanNet}&\multirow{2}{*}{GFLOPs}&\multirow{2}{*}{\#Params}\\
&mIoU&mAcc&OA&mIoU&mAcc&OA&mIoU&&\\
\midrule
KPConv~\cite{kpconv}&70.6&79.1&-&67.1&72.8&-&68.4&-&-\\
PointTransformer~\cite{zhao2021point}&73.5&81.9&90.2&70.4&76.5&90.8&70.6&2.80&-\\
RepSurf~\cite{repsurf}&74.3&82.6&90.8&68.9&76.0&90.2&70.0&1.04&-\\
RandLA~\cite{hu2020randla}&70.0&81.5&88.0&-&-&-&-&5.8&-\\
PTV2~\cite{wu2022point}&-&-&-&71.6&77.9&91.1&75.4&-&-\\
PointNeXt-XL~\cite{pointnetxt}&74.9&83.0&90.3&71.1&77.2&91.0&71.5&84.8&41\\
PointVector-XL~\cite{pointvector}&78.4&86.1&91.9&72.3&78.1&91&-&58.5&24.1\\
\midrule
PointNet++~\cite{pointnet++}&54.5&67.1&81.0&53.5&-&83.0&-&-&-\\
+\textbf{X-3D}&~~~~~~~62.9\textcolor{green!40!gray}{\small $\uparrow$7.4} &~~~~~~~78.3\textcolor{green!40!gray}{\small $\uparrow$11.2}
&~~~~~~~84.4\textcolor{green!40!gray}{\small $\uparrow$1.1}&~~~~~~~57.8\textcolor{green!40!gray}{\small $\uparrow$4.3}&-&~~~~~~~85.3\textcolor{green!40!gray}{\small $\uparrow$2.3}&-&-&-\\
\midrule
PointMetaBase-L~\cite{lin2023meta}&75.6&-&90.6&69.7&-&90.7&71.0&2.0&2.7\\
+\textbf{X-3D}&~~~~~~~76.7\textcolor{green!40!gray}{\small $\uparrow$1.1}&-&~~~~~~~91.1\textcolor{green!40!gray}{\small $\uparrow$0.5}
&~~~~~~~71.9\textcolor{green!40!gray}{\small $\uparrow$2.2}&-&~~~~~~~91.2\textcolor{green!40!gray}{\small $\uparrow$0.6}&~~~~~~~71.8\textcolor{green!40!gray}{\small $\uparrow$0.8}&2.2&3.8\\
PointMetaBase-XL~\cite{lin2023meta}&76.3&-&91.0&71.6&-&90.6&71.8&9.2&15.3\\
+\textbf{X-3D}&~~~~~~~77.7\textcolor{green!40!gray}{\small $\uparrow$1.4}&-&~~~~~~~91.6\textcolor{green!40!gray}{\small $\uparrow$0.6}&~~~~~~~72.1\textcolor{green!40!gray}{\small $\uparrow$0.5}&-&~~~~~~~91.4\textcolor{green!40!gray}{\small $\uparrow$0.8}&~~~~~~~72.8\textcolor{green!40!gray}{\small $\uparrow$1.0}&9.8&18.1\\
\midrule
DeLA~\cite{dela}&78.3&86.1&91.7&74.1&80.0&92.2&75.9&14.0&7.0\\
+\textbf{X-3D}&~~~~~~~\textbf{79.2}\textcolor{green!40!gray}{\small $\uparrow$0.9}&~~~~~~~\textbf{86.5}\textcolor{green!40!gray}{\small $\uparrow$0.4}&~~~~~~~\textbf{91.9}\textcolor{green!40!gray}{\small $\uparrow$0.2}&~~~~~~~\textbf{74.3}\textcolor{green!40!gray}{\small $\uparrow$0.2}&~~~~~~~\textbf{80.1}\textcolor{green!40!gray}{\small $\uparrow$0.1}&\textbf{92.2}&~~~~~~~\textbf{76.3}\textcolor{green!40!gray}{\small $\uparrow$0.4}&14.8&8.0\\
\bottomrule
\end{tabular}
}
\vspace{1mm}
\end{table*}

Due to the randomness of kNN and ball query, the constructed neighborhood usually produces overlapping parts. Therefore, for the center point $\mathbf{p}_{i}$ of a neighborhood, it may also be a neighborhood point $\mathbf{p}_{*,i}$ of other neighborhoods at the same time. We use this overlap phenomenon to fuse the features of the $i$-th point in different situations to propagate the context of different neighborhoods. If the $i$-th point is a center point, its feature $\hat{\mathbf{f}}_i$ is obtained as follows:
\begin{equation}
    \hat{\mathbf{f}}_i=MaxPool(\hat{\mathbf{f}}_{i,j}) ,j=1,2...k
\end{equation}
Then if the $i$-th point is a neighborhood point in the neighborhood centered at $j$-th point, its feature $\hat{\mathbf{f}}_{j,i}$ is obtained from (\ref{eq:update}). However, a point may be simultaneously captured as a neighborhood point by more than one neighborhood, so when the $i$-th point is a neighborhood point, we define the set of neighborhoods containing $i$-th point as $\mathcal{G}_i$. So for each element $j$ in $\mathcal{G}_i$, there will be $i\in \mathcal{N}(\mathbf{p}_j)$, where $\mathcal{N}(\mathbf{p}_j)$ represents the neighborhood centered at $j$-th point. Thus we will aggregate the features at $i$-th point  in $\mathcal{G}$ representing the local structural context:
\begin{equation}
    \hat{\mathbf{f}}^{(1)}_i=\frac{1}{\mathop{Len}(\mathcal{G}_i)}\sum_j \hat{\mathbf{f}}_{j,i}, j\in \mathcal{G}_i
\end{equation}

Finally, we dynamically fuse the current neighborhood features with the context:
\begin{equation}
    \hat{\mathbf{f}}^{(2)}_i=MLP([\hat{\mathbf{f}}^{(1)}_i,\hat{\mathbf{f}}_i])
\end{equation}



\vspace{-0.7em}
\section{Experiments}

\begin{table}[!t]
\centering
\caption{Classification results on the ScanObjectNN dataset. X-3D only adds an additional 0.1 M and 0.1 GFLOPs for a 0.7\% OA improvement, achieving new state-of-the-art performance (\textbf{rank1 without extra training data}).
}
\label{tab:classication}
\resizebox{\linewidth}{!}{
\begin{tabular}{c|cc|cc}
\toprule
\multirow{2}{*}{Method}&\multicolumn{2}{c|}{\textbf{ScanObjectNN}}&\multirow{2}{*}{GFLOPs}&\multirow{2}{*}{\#Params}\\
&OA&mAcc&\\
\midrule
DGCNN~\cite{dgcnn}&78.1&73.6&2.43&1.82M\\
PointNet~\cite{pointnet}&68.2&63.4&0.45&3.47M\\
PointNet++~\cite{pointnet++}&77.9&75.4&1.7&1.5M\\
MVTN~\cite{mvtn}&82.8&-&1.78&4.24M\\
SPoTr~\cite{sptr}&88.6&86.8&-&-\\
PointMLP~\cite{pointmlp}&85.7&84.4&31.4&13.2M\\
PointNeXt-S~\cite{pointnetxt}&88.20&86.84&1.64&1.4M\\
PointVector~\cite{pointvector}&88.17&86.69&-&1.55M\\
\midrule
PointMetaBase-S~\cite{lin2023meta}&88.2&86.90&0.55&1.4M\\
+\textbf{X-3D}&~~~~~~~88.83\textcolor{green!40!gray}{\small $\uparrow$0.63}&~~~~~~~87.16\textcolor{green!40!gray}{\small $\uparrow$0.26}&0.59&1.5M\\
\midrule
DeLA~\cite{dela}&90.4&89.3&1.50&5.3M\\
+\textbf{X-3D}&~~~~~~~\textbf{90.7}\textcolor{green!40!gray}{\small $\uparrow$0.3}&~~~~~~~\textbf{89.9}\textcolor{green!40!gray}{\small $\uparrow$0.6}&1.53&5.4M\\

\bottomrule
\end{tabular}
}

\vspace{-1.1em}

\end{table}

We evaluated X-3D on four tasks: segmentation, classification, object part segmentation, and detection to prove its effectiveness. At the same time, we embedded X-3D on the simplest and most advanced models to demonstrate its applicability.  Further, we designed a series of ablation experiments to demonstrate the effectiveness of various parts of X-3D. In the supplementary material, there will be more detailed experiments .

\begin{table*}[!t]
\center
\caption{Detection results on the ScanNetV2~\cite{scannet} and the SUN RGB-D~\cite{sunrgb-d} dataset. We tested the performance of X-3D on GroupFree~\cite{groupfree} and report on mAP@0.25 and mAP@0.50.
}
\label{tab:detection}
\resizebox{0.9\linewidth}{!}{
\begin{tabular}{c|c|cc|cc|cc}
\toprule
\multirow{2}{*}{Method}&\multirow{2}{*}{BackBone}&\multicolumn{2}{c|}{\textbf{ScanNetV2}}&\multicolumn{2}{c|}{\textbf{SUN RGB-D}}&\multirow{2}{*}{\#Param}&\multirow{2}{*}{GFLOPs}\\
&&mAP@0.25&mAP@0.50&mAP@0.25&mAP@0.50\\
\midrule
VoteNet~\cite{votenet}&PointNet++&62.9&39.9&59.1&35.8&-&-\\
ImVoteNet~\cite{imvotenet}&PointNet++&-&-&63.4&-&-\\
H3DNet~\cite{h3dnet}&4$\times$ PointNet++&67.2&48.1&60.1&39.0&-&-\\
3DETR~\cite{3ddetr}&Transformer&65.0&47.0&59.1&32.7&-&-\\
BRNet~\cite{brnet}&PointNet++&66.1&50.9&61.1&43.7&-&-\\
\midrule
GroupFree$^{6,256}$&PointNet++&67.3&48.9&63.0&45.2&11.49&6.7\\
GroupFree$^{6,256}$&RepSurf-U&68.8&50.5&\textbf{64.3}&45.9&11.50&6.8\\
GroupFree$^{6,256}$&\textbf{X-3D}&~~~~~~~\textbf{69.0}\textcolor{green!40!gray}{\small $\uparrow$1.7}&~~~~~~~\textbf{51.1}\textcolor{green!40!gray}{\small $\uparrow$2.2}&~~~~~~~\textbf{64.5}\textcolor{green!40!gray}{\small $\uparrow$1.5}&~~~~~~~\textbf{46.9}\textcolor{green!40!gray}{\small $\uparrow$1.7}&12.30&7.1\\
\bottomrule
\end{tabular}
}

\vspace{1mm}

\end{table*}

\subsection{Segmentation}
We evaluated X-3D on two datasets for scene segmentation, S3DIS~\cite{s3dis} and ScanNet~\cite{scannet}.

\begin{table}[!t]
\centering
\caption{ 3D Outdoor Object Detection Results on KITTI dataset on 3DSSD. We report the mIoU at three levels of difficulty }
\label{tab:kitti}

\resizebox{0.9\linewidth}{!}{
 \setlength{\tabcolsep}{2.5pt}
\begin{tabular}{c|cccc}
\toprule
Method&easy&moderate&hard\\
\midrule
3DSSD+\textbf{X-3D}&~~~~87.6(\textbf{88.5})&~~~~78.1(\textbf{78.3})&~~~~76.7(\textbf{76.9})\\
\bottomrule
\end{tabular}

}
\end{table}

\textbf{S3DIS.} There are six large indoor scenes, which contain 271 rooms and 13 categories in S3DIS. Based on the generic setting, we evaluated X-3D on 6-Fold and Area5 and reported the mIoU, mAcc and OA.  As shown in Table~\ref{tab:segmentation}, X-3D can effectively improve its performance on different scale models such as PointNet++~\cite{pointnet++}, PointMetaBase~\cite{lin2023meta}, DeLA~\cite{dela}, and the extra computational overhead is acceptable. For PointNet++, X-3D improves 4.3\% and 7.4\% mIoU on Area5 and 6-Fold, respectively. For PointMetaBase, We evaluated X-3D on its lightweight version PointMetaBase-L as well as its large-scale version PointMetaBase-XL. Although PointMetaBase-L is small in scale, it only brings 1.1 M and 0.2 GFLOPs extra overhead by embedding X-3D and improves 2.2\% and 1.1\% mIoU on Area 5 and 6-Fold, exceeding the XL version.  And for PointMetaBase-XL, X-3D improves 0.5\% and 1.4\% mIoU on Area5 and 6-Fold. Finally, we tested X-3D on DeLA, which improves 0.2\% and 0.9\% mIoU on Area 5 and 6-Fold and achieves the new state-of-the-art performance.

\textbf{ScanNet}. There are 1201 indoor training point clouds and 312 validation point clouds in  ScanNet. We evaluate X-3D on PointMetaBase-L, PointMetaBase-XL and DeLA, which boost 0.8\%,1.0\%,0.4\% mIoU respectively. 

Through further analysis, we found that X-3D performs well on more complex datasets, such as S3DIS 6-Fold and ScanNet and simpler models such as PointNet++ and PointMetaBase-L, because it explicitly introduces an effective geometric prior, which greatly reduces the difficulty of model learning.

\subsection{Classification}
We evaluated X-3D on PointMetaBase-S~\cite{lin2023meta} and DeLA~\cite{dela} on ScanObjectNN~\cite{scanobjectnn} and reported mAcc and OA, which is shown in Table~\ref{tab:classication}. There are 15,000 actual scanned objects divided into 15 classes with 2,902 unique object instances in ScanObjectNN. More importantly, since ScanObjectNN is a real-world dataset, it introduces background, noise and other factors, which increases the difficulty of model learning. Due to the efficiency of X-3D, adding only an additional 0.04 GFLOPs and 0.1 M parameters leads to a 0.7\% OA improvement on PointMetaBase and 0.3\%OA improvement on DeLA with only 0.03 GLLOPs and 0.1M parameters, achieving new state-of-the-art performance.

\begin{table}[!t]
\centering
\caption{ Part segmentation results on the ShapeNetPart~\cite{shapenetpart} dataset. We tested X-3D on the state-of-the-art model PointMetaBase~\cite{lin2023meta}, which can effectively improve the performance.
}
\label{tab:partseg}
\resizebox{\linewidth}{!}{
\begin{tabular}{c|cc|cc}
\toprule
Method&ins. mIoU&cls. mIoU&Params&GFLOPs\\
\midrule
PointNet++~\cite{pointnet++}&85.1&81.9&1.0&4.9\\
KPConv~\cite{kpconv}&86.4&85.1&-&-\\
PointTransformer~\cite{zhao2021point}&86.6&83.7&7.8&-\\
PointMLP~\cite{pointmlp}&86.1&85.1&-&-\\
StratifiedFormer~\cite{lai2022stratified}&86.6&85.1&-&-\\
PointNeXt-S~\cite{pointnetxt}&86.7&85.6&1.0&4.5\\
PointMetaBase-S (C=64)~\cite{lin2023meta}&86.9&85.1&3.8&3.85\\
\midrule
PointMetaBase-S (C=32)~\cite{lin2023meta}&86.7&84.4&1.0&1.39\\
\textbf{+X-3D}&~~~~~~~\textbf{87.0}\textcolor{green!40!gray}{\small $\uparrow$0.3}&~~~~~~~\textbf{85.1}\textcolor{green!40!gray}{\small $\uparrow$0.7}&2.0&2.0\\
\bottomrule
\end{tabular}
}

\vspace{1mm}

\end{table}

\subsection{Indoor Detection}
We evaluated X-3D on GroupFree$^{6,256}$~\cite{groupfree} on two datasets ScanNetV2~\cite{scannet} and SUN RGB-D~\cite{sunrgb-d} in Table~\ref{tab:detection}. ScanNetV2 contains 1513 indoor scenes and 18 object classes. SUN RGB-D contains 5K indoor RGB and depth images. Based on the generic setting, we report the mean Average Precision under the thresholds of 0.25  and 0.5.  For ScanNetV2, X-3D improves 1.7\%mAP@0.25 and 2.2\% mAP@0.5. For SUN RGB-D, X-3D improves 1.5\% mAP@0.25 and 1.7\% mAP@0.50.

\subsection{Outdoor Detection}
As shown in Table~\ref{tab:kitti}, we applied X-3D to 3DSSD~\cite{3dssd} and tested it on the KITTI~\cite{kitti} validation dataset. X-3D improves 0.9\%,0.2\%,0.2\% mIoU on easy, moderate, and hard, respectively.

\subsection{Part Segmentation}
We evaluated X-3D on the state-of-the-art model PointMetaBase-S~\cite{lin2023meta} on the ShapeNetPart~\cite{shapenetpart} dataset, where the results are shown  in Table~\ref{tab:partseg}. There are 16880 models with 16 different shape categories and 50 part labels in ShapeNetPart.  Based on the generic setting, we report the final performance using the voting strategy. For PointMetaBase-S, X-3D improves 0.3\% ins. mIoU and 0.7\%cls. mIoU.

\begin{figure}[!t]
    \centering
    \begin{subfigure}[t]{0.4\linewidth}
           \centering
           \includegraphics[width=0.75\linewidth]{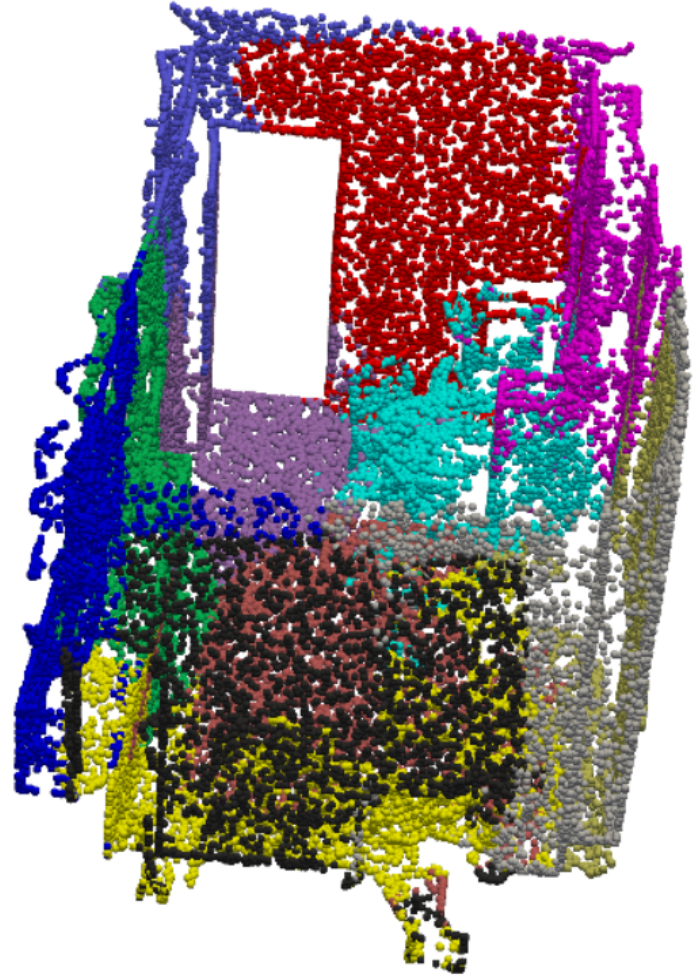}
           \caption{Scene Structure Kernel}
           \label{fig: vis_kernel_scene}
           
    \end{subfigure}
    \begin{subfigure}[t]{0.4\linewidth}
            \centering
            \includegraphics[width=0.9\linewidth]{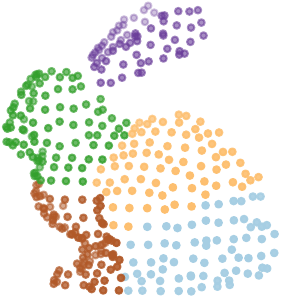}
            \caption{Object Structure Kernel} 
            \label{fig: vis_kernel_object}
    \end{subfigure}
  
    \caption{Kmeans Results of the Structure Kernel. The parameters of structure kerne are restricted by local structure, which provides a good structural prior. }
    \label{fig:vis_kernel}
    \vspace{-1.1em}
\end{figure}

\subsection{Analysis}
In this section, we analyze in detail why X-3D works and compare it with Transformer.

\textbf{Why X-3D works?} 
We believe that the essence of X-3D effectiveness is similar to the principle of LLE methods in manifold learning.
 First, LLE limits the parameters of the mapping through the local structure (please refer to the supplementary material for details). Similarly, we argue that X-3D restricts the kernel parameter by its explicit structure, which greatly reduces the difficulty of model learning.  To prove the validity of the restriction, we visualize the distribution of structure kernels and classify kernels with k-means in Figure~\ref{fig:vis_kernel}. 
We find that the structure kernel is closely related to the geometric structure, such as the \textcolor{Purple}{ears}, \textcolor{Green}{head}, \textcolor{Brown}{feet} and \textcolor{Blue}{tail} of the object in Figure~\ref{fig: vis_kernel_object}, as well as the walls, chairs and floors in the scene in Figure~\ref{fig: vis_kernel_scene}. Therefore, the structure kernel contains explicit structural information to provide effective geometric priors for the model.

\begin{table}[!t]
\centering
\caption{We embed X-3D into Point Transformer by generating relational position embedding with our X-3D module and report the mIoU on S3DIS and ScanNet   }
\label{tab:trans_performance}
\vspace{-0.7em}
\resizebox{0.7\linewidth}{!}{
\begin{tabular}{c|c|c}
\toprule
&S3DIS&ScanNet\\
\hline
Point Transformer&70.4&70.6\\
+X-3D&71.6~~~(\textcolor{green}{+1.2})&73.9~~~(\textcolor{green}{+3.3})\\
\bottomrule
\end{tabular}
}
\vspace{0.6em}
\end{table}

\textbf{Comparison with Transformer.}
Firstly, there are some differences between Point Transformer~\cite{zhao2021point,lai2022stratified} and implict high-dimensional structure modeling. Although the transformer dynamically generates the  kernel corresponding to each relation vector, it considers local structural information during the generation process. For example, the process of PointTransformer~\cite{zhao2021point} can be written as:
\begin{equation}
    M_{i,j}=MLP(\mathbf{f}_i-\mathbf{f}_{i,j})+MLP(\mathbf{p}_i-\mathbf{p}_{i,j})
\end{equation}
\begin{equation}
   IS_i=\sum_{j=1}^kexp(M_{i,j})
\end{equation}
\begin{equation}
\label{eq:trans}
   \mathbf{W}_{i,j}=\theta(M_{i,j},IS_i)
\end{equation}
Compare to the (\ref{eq:trans}) with (\ref{eq:ls_structure}), the main difference between Transformer and X-3D is that we use the explicit structure $ES_i$ directly captured in the data while Transformer uses the implicit structure $IS_i$ learned by the network.  We argue that the explicit local structure greatly reduces the difficulty of model learning. In Table~\ref{tab:segmentation}, we have performed a performance comparison of X-3D with PointTransformer. In this case, we argue that X-3D is fundamentally different from Transformers and can be applied to Transformers as well. Therefore, we embedded X-3D into Point Transformer by generating relational position embedding with our X-3D module instead of the original linear projection and the results are reported in Table~\ref{tab:trans_performance}.

\begin{table}[!t]
\centering
\caption{We compare computational cost  between X-3D and Transformer. }
\label{tab:trans_time}
\vspace{-0.7em}
\resizebox{0.85\linewidth}{!}{
\begin{tabular}{c|c|c|c}
\toprule
&{c=256,k=16}&{c=256,k=32}&{c=512,k=16}\\
\midrule
X-3D (structure kernel)&0.24&0.34&0.46\\
PTv1 (vector attention)&1.51&2.21&5.83\\
VIT (scalar attention)&0.84&1.02&3.29\\
\bottomrule
\end{tabular}
}

\end{table}

More importantly, compared with transformer-generating vector kernels for a single vector, X-3D generates a shared dynamic structure kernel for all vectors in a local neighborhood, which reduces a significant amount of computational cost. Here we provide the comparison between the computational cost in further detail. For an input of shape $N*K*(C+3)$, where $N$ is the number of points, $C$ is the feature channel number, and $K$ is the number of neighborhood points. We set different $C$ and $K$ and report the results (GFLOPs). From Table~\ref{tab:trans_time}, we see that X-3D can save a lot of computation costs.

\begin{table}[!t]

\centering
\caption{We tested the effect of different utilization of $ES_i$. We report the Euclidean distance (GAP) between the local structure of the
embedding space and the input space and the OA. }
\label{tab:kernel}

\resizebox{0.8\linewidth}{!}{
 \setlength{\tabcolsep}{2.5pt}
\begin{tabular}{c|cccc}
\toprule
&GAP&OA&\#Params&GFLOPs\\
\midrule
baseline&5.3&88.2&1.4&0.55\\
Concat&3.1&88.3&1.4&0.58\\
Vector Kernel&1.6&88.6&1.5&0.83\\
\midrule
Structure Kernel&\textbf{1.3}&\textbf{88.8}&1.5&0.59\\
\bottomrule
\end{tabular}
}
\vspace{-1.1em}
\end{table}
\vspace{1.1em}

\subsection{Ablation Study}
\textbf{Structure Kernel.} As shown in Table~\ref{tab:kernel}, we tested the effect of different utilization of $ES_i$ on ScanObjectNN.  In the first row we report the performance of the original model. Then in the second row, we try to directly merge the explicit local structure with the relation vector. Although the explicit introduction of the local structure information reduces the GAP, the simple merging cannot model the relation vector through the strong restriction of the local structure, resulting in limited performance improvement. Further, in the third row, we try to generate vector kernels according to the correlation between each relation vector and the explicit local structure which is similar to transformer. Although this method strongly restricts the modeling of relation vectors by the explicit local structure and has higher performance, it also brings a large computational cost. Finally, in the last row, we report the performance of structure kernel, since it directly generates a shared kernel for all relation vectors according to the local structure, the computational cost and performance is greatly reduced 
and improved.

\textbf{Denoising and Neighborhood Context Propagation.} As shown in Table~\ref{tab:neighborhood}, we test the effect of denoising step and neighborhood context propagation step on PointMetaBase-L on S3DIS. For denoising step, noise points may distort local structural information, which is also discussed in RepSurf and other methods, so the denoising process is important, and we see that it can improve performance on various datasets and tasks. For neighborhood context propagation step, its purpose is to solve the problem caused by randomly selecting neighborhood points. However, for the classification task, the number of points is small, so the influence of neighborhood randomness is small. It can be seen that this step has little improvement for the classification task, but has a great performance improvement for the segmentation task with a large number of points.


\begin{table}[!t]
\centering
\caption{We test the effect of the denoising step (\textbf{DN}) and the neighborhood context propagation (\textbf{NCP}) step on PointMetaBase on S3DIS Area5 and ScanObjectNN. }
\label{tab:neighborhood}

\resizebox{\linewidth}{!}{
 \setlength{\tabcolsep}{2.5pt}
\begin{tabular}{c|cccc}
\toprule
&&+ DN&+ NCP& + DN\&NCP\\
\midrule
S3DIS (mIoU\%)&70.9&71.3&71.4&71.9\\
ScanObjectNN (OA\%)&88.5&88.8&88.6&88.8\\
\bottomrule
\end{tabular}
}
\vspace{0.6em}
\end{table}

\textbf{Robustness.} 
Since X-3D inherits the advantages of LLE and is more robust to local transformation, 
we report the performance of different models on ScanObjectNN before and after transformation in Table~\ref{tab:rotation}.  We see that X-3D is more robust because the introduction of the explicit structure strengthens the feature extraction ability of the model for different structures.

\begin{table}[!t]
\centering
\vspace{-0.7em}
\caption{We test the robustness of X-3D on ScanObjectNN with different transformations. }
\label{tab:rotation}
\resizebox{\linewidth}{!}{
\begin{tabular}{c|c|ccc|cc}
\toprule
&\multirow{2}{*}{Vanilla}&\multicolumn{3}{c|}{Rotation}&\multicolumn{2}{c}{Scale}\\
&&30&60&90&0.9&1.1\\
\midrule
X-3D &88.8&88.6~~~(\textcolor{green}{-0.2})&88.3~~~(\textcolor{green}{-0.5})&88.5~~~(\textcolor{green}{-0.3})&88.6~~~(\textcolor{green}{-0.2})&88.7~~~(\textcolor{green}{-0.1})\\
PointMetaBase&88.2&87.4~~~(\textcolor{green}{-0.8})&87.1~~~(\textcolor{green}{-1.1})&87.7~~~(\textcolor{green}{-0.5})&87.4~~~(\textcolor{green}{-0.8})&87.5~~~(\textcolor{green}{-0.7})\\

\bottomrule
\end{tabular}
}
\vspace{-1.1em}
\end{table}

\subsection{Visualization}
\begin{figure}[!h]
 \begin{subfigure}[t]{0.12\textwidth}
           \centering
           \includegraphics[width=\textwidth]{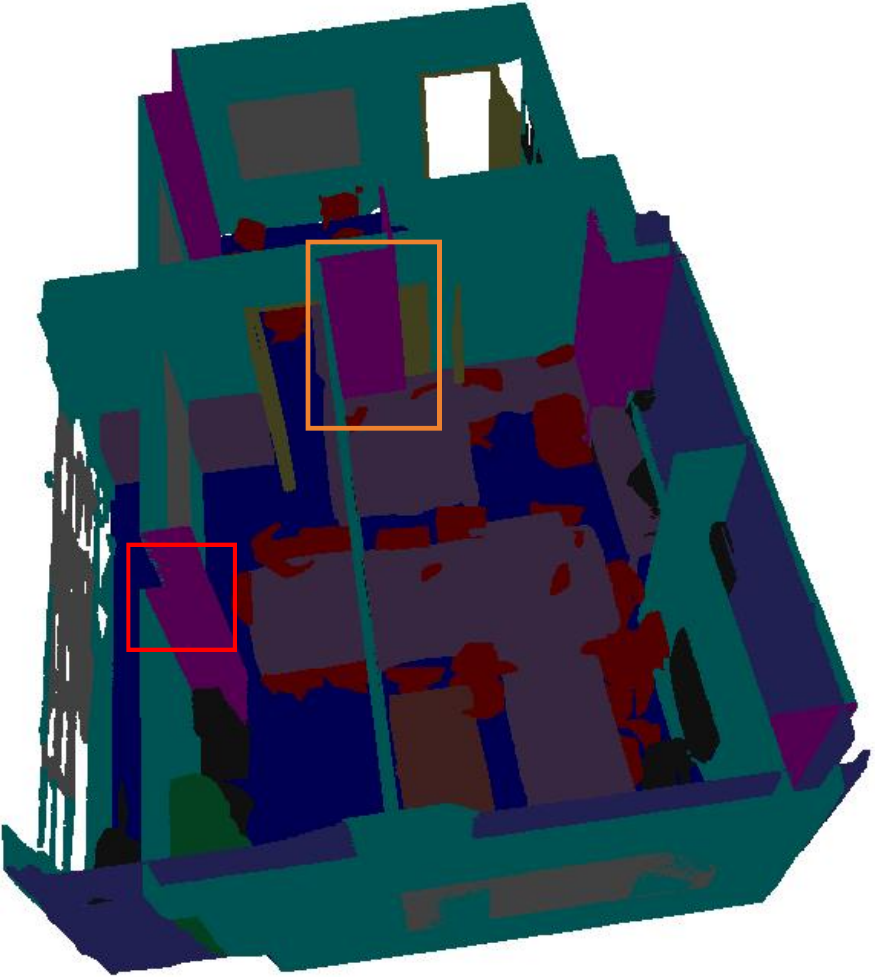}
            \caption{Ground Truth}
           
\end{subfigure}
\begin{subfigure}[t]{0.13\textwidth}
           \centering
           \includegraphics[width=\textwidth]{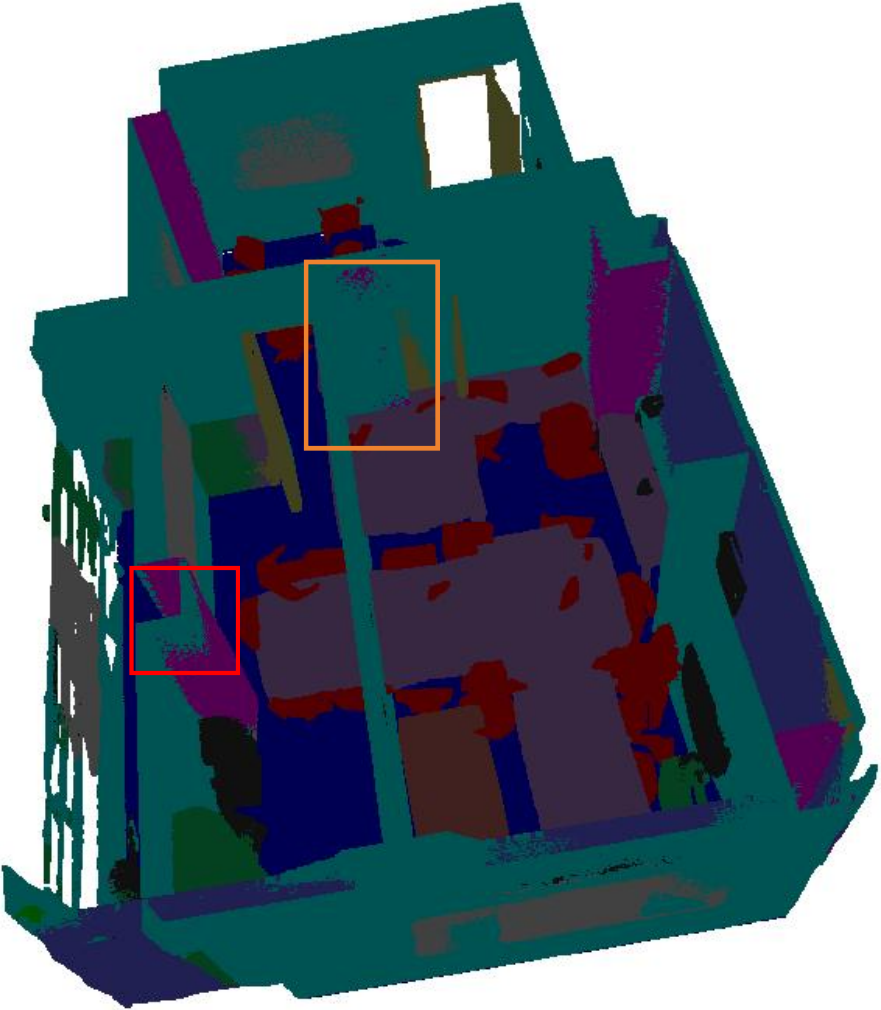}
            \caption{PointMetaBase}
            
\end{subfigure}
\begin{subfigure}[t]{0.12\textwidth}
           \centering
           \includegraphics[width=\textwidth]{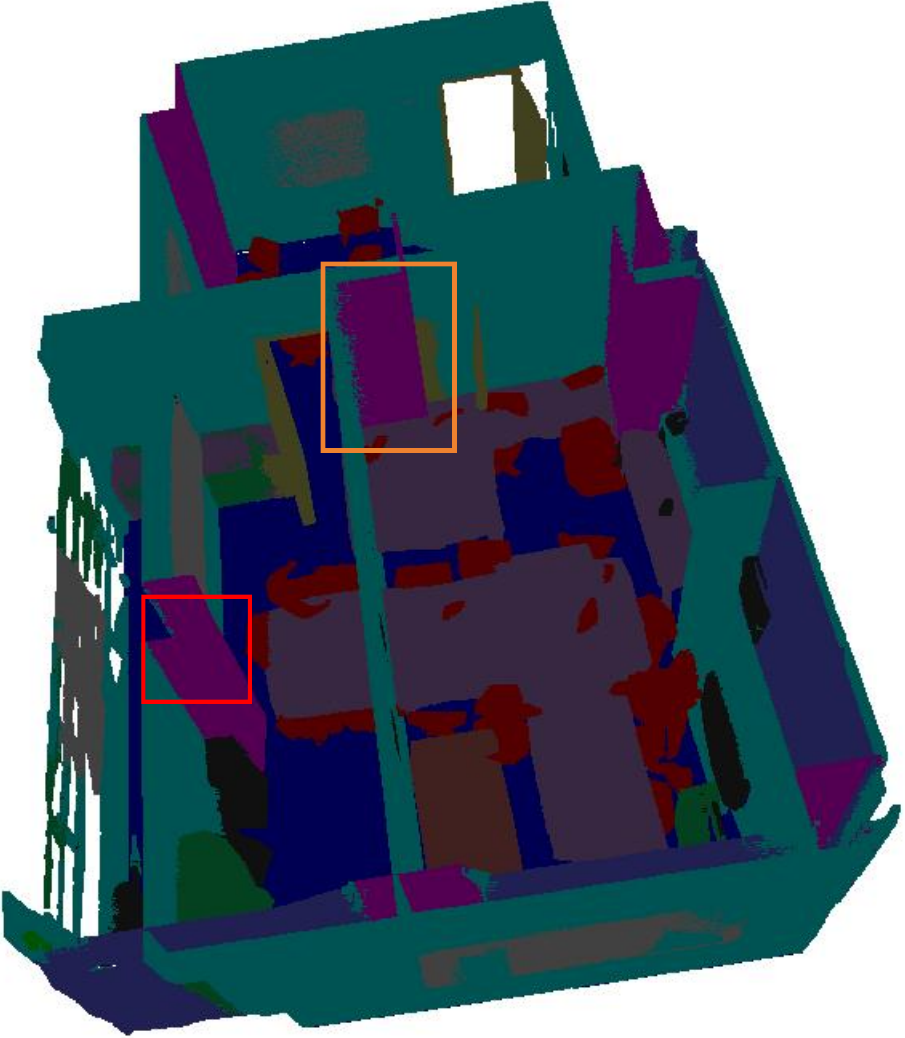}
            \caption{+X-3D}
            
\end{subfigure}
\begin{subfigure}[t]{0.15\textwidth}
        \centering
        \includegraphics[width=\textwidth]{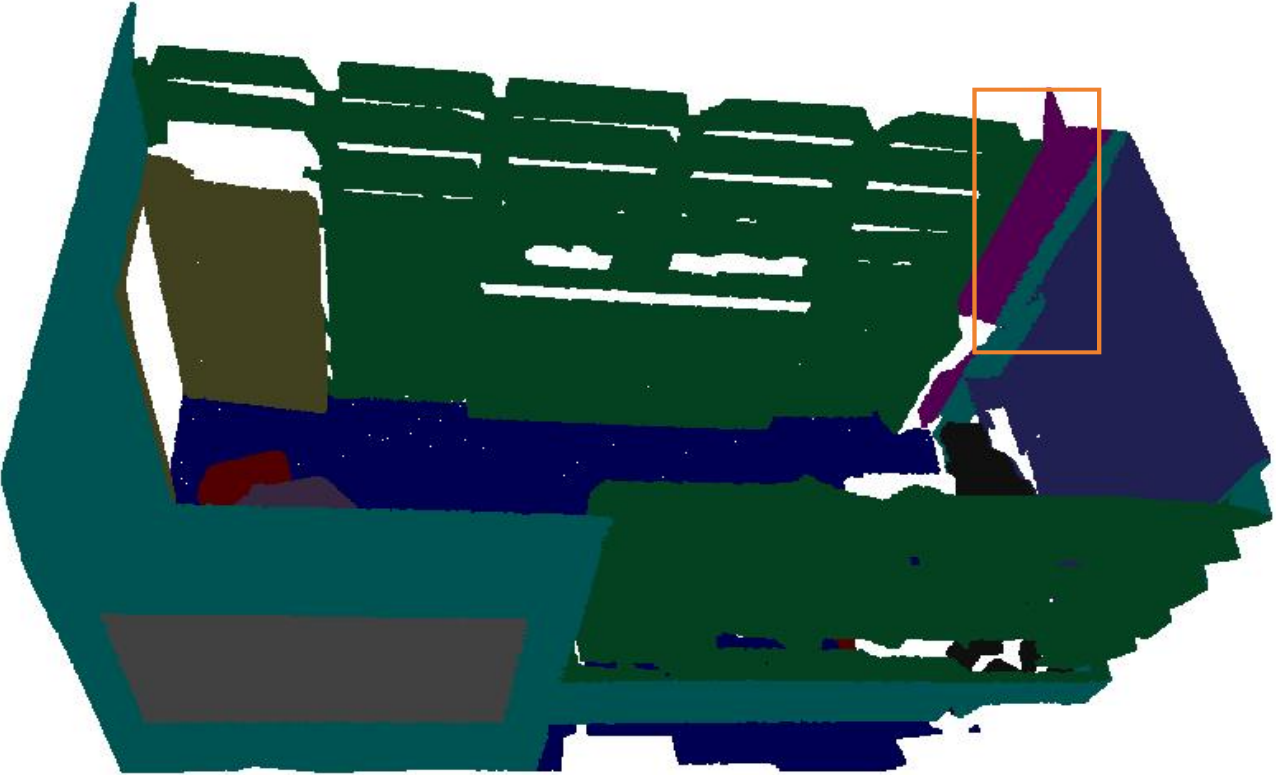}

       \caption{Ground Truth}
        
\end{subfigure}
\begin{subfigure}[t]{0.15\textwidth}
           \centering
           \includegraphics[width=\textwidth]{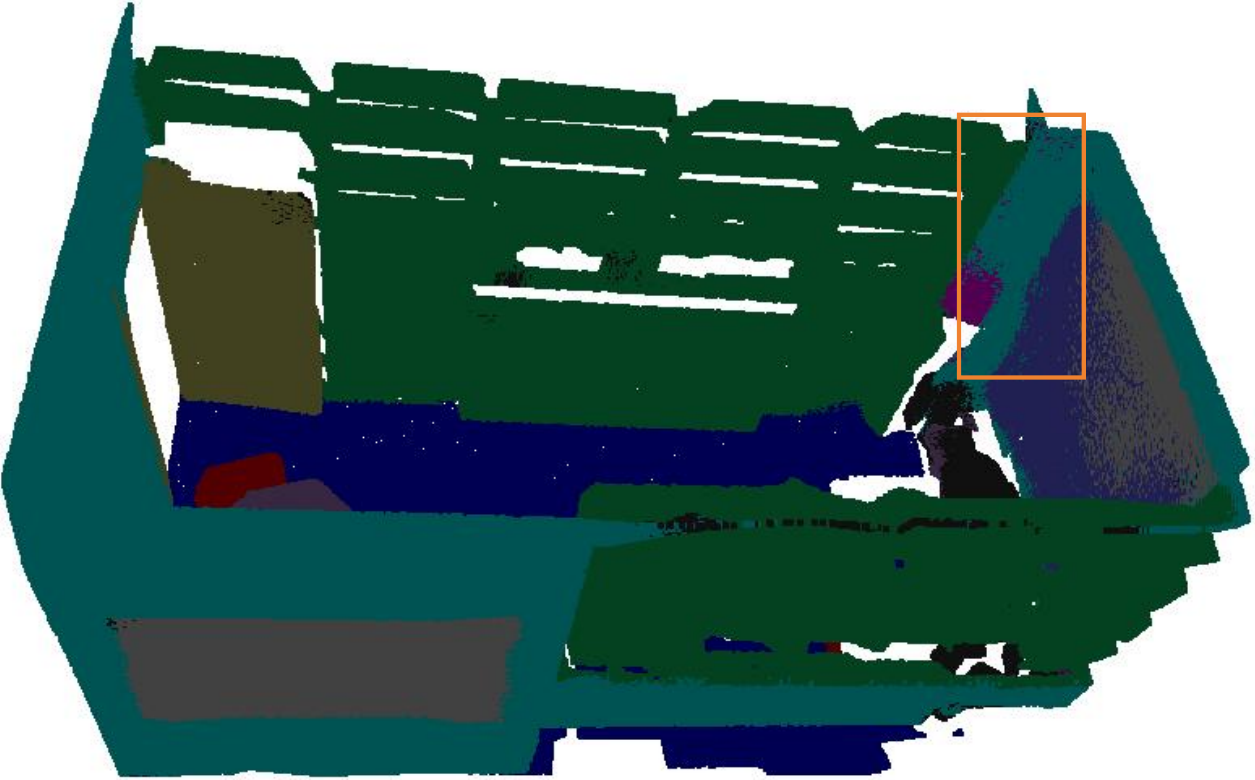}
            \caption{PointMetaBase}
            
\end{subfigure}
\begin{subfigure}[t]{0.15\textwidth}
           \centering
           \includegraphics[width=\textwidth]{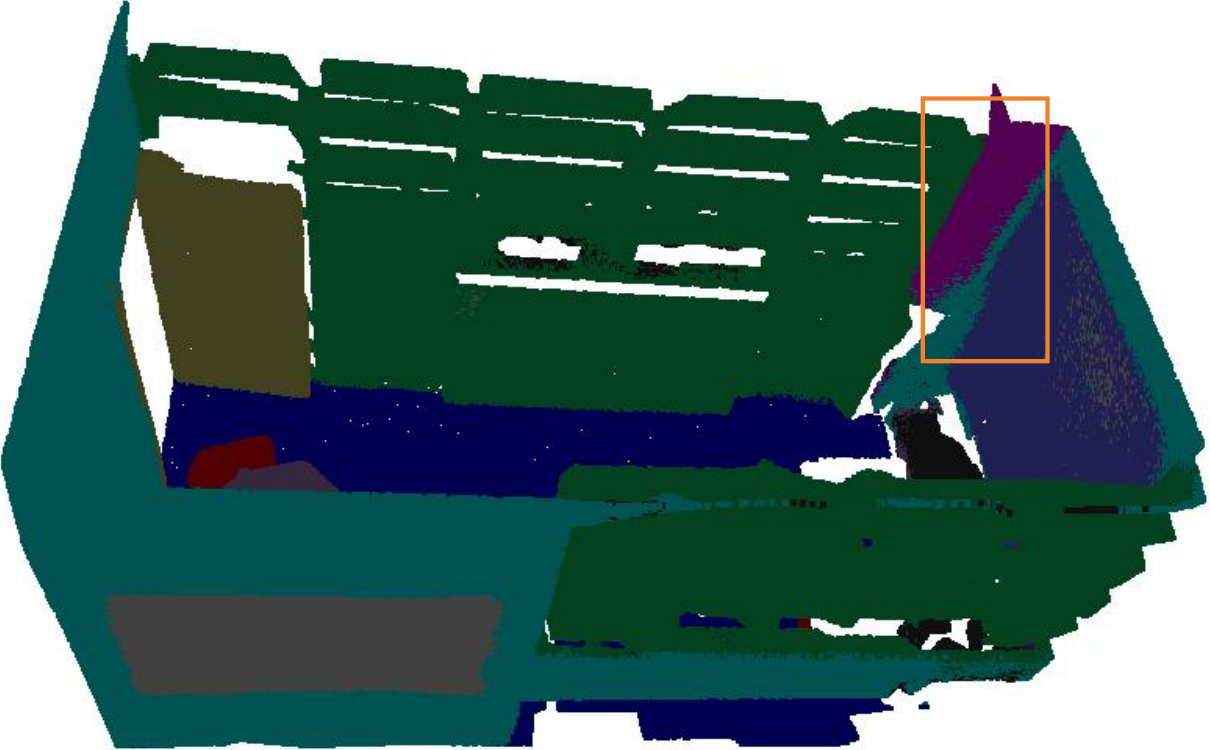}
             \caption{+X-3D}
           
\end{subfigure}
\caption{We visualize the segmentation results of PointMetaBase-L as well as X-3D on S3DIS Area5 and annotate the different places
}\label{fig:vis_result}
\vspace{-1.1em}
\end{figure}

As shown in Fig~\ref{fig:vis_result}, we visualize the segmentation results on S3DIS and can see that X-3D segmentation prediction is generally more complete and accurate due to the introduction of explicit local structure information.

\section{Conclusion}  In this paper, we present X-3D, an  explicit 3D structure-based modeling approach which uses an explicit input structure to limit the embedding parameters. Experimental results show that X3D can effectively reduce the difficulty of model learning and achieve the latest state-of-the-art performance

\textbf{Acknowledgements.}
This work was supported in part by the National Natural Science Foundation of China under Grant 62376032 and Grant U22B2050.

{
    \small
    \bibliographystyle{ieeenat_fullname}
    \bibliography{main}
}

\end{document}